\documentclass[conference]{IEEEtran}
\usepackage{times}

\usepackage[numbers]{natbib}
\usepackage{multicol}
\usepackage[bookmarks=true]{hyperref}
\usepackage[english]{babel}
\usepackage[final]{graphicx}
\usepackage{amsmath}
\usepackage{amssymb} 
\usepackage{siunitx}
\usepackage{nicefrac}
\usepackage{amsmath}
\usepackage{amssymb}
\usepackage{bm}
\usepackage{color}
\usepackage{graphicx}

\begin{document}

\title{Contact-Implicit Optimization of Locomotion Trajectories for a Quadrupedal Microrobot}

\author{\authorblockN{Neel~Doshi$^{1,2}$, Kaushik~Jayaram$^{1,2}$, Benjamin~Goldberg$^{1,2}$,  Zachary~Manchester$^3$, \\  Robert~J.~Wood$^{1,2}$, and Scott~Kuindersma$^1$}
\authorblockA{$^1$John A. Paulson School of Engineering and Applied Sciences, Harvard University, Cambridge, M.A \\ Email: ndoshi@g.harvard.edu, scottk@seas.harvard.edu}
\authorblockA{$^2$Wyss Institute for Biologically Inspired Engineering, Harvard University, Cambridge, M.A.}
\authorblockA{$^3$Department of Aeronautics and Astronautics, Stanford University, Stanford, C.A.}}

\maketitle

\begin{abstract}
		Planning locomotion trajectories for legged microrobots is challenging because of their complex morphology, high frequency passive dynamics, and discontinuous contact interactions with their environment. Consequently, such research is often driven by time-consuming experimental methods. As an alternative, we present a framework for systematically modeling, planning, and controlling legged microrobots. We develop a three-dimensional dynamic model of a 1.5\,g quadrupedal microrobot with complexity (e.g., number of degrees of freedom) similar to larger-scale legged robots. We then adapt a recently developed variational contact-implicit trajectory optimization method to generate feasible whole-body locomotion plans for this microrobot, and we demonstrate that these plans can be tracked with simple joint-space controllers. We plan and execute periodic gaits at multiple stride frequencies and on various surfaces.  These gaits achieve high per-cycle velocities, including a maximum of 10.87\,mm/cycle, which is 15\% faster than previously measured velocities for this microrobot. Furthermore, we plan and execute a vertical jump of 9.96\,mm, which is 78\% of the microrobot's center-of-mass height. To the best of our knowledge, this is the first end-to-end demonstration of planning and tracking whole-body dynamic locomotion on a millimeter-scale legged microrobot. 
\end{abstract}
\section{INTRODUCTION}

\subsection{Motivation}

Laminate manufacturing processes, such as smart-composite microstructures \cite{wood2008microrobot} and printed-circuit MEMS \cite{whitney2011pop} enable rapid and reliable assembly of flexure-based, millimeter-scale devices. One major advantage of these manufacturing techniques is the ability to realize mechanically complex devices with many degrees-of-freedom (DOFs) at the millimeter scale. This has enabled the development of small scale legged microrobots that do not sacrifice dexterity. These legged microrobots leverage favorable inertial scaling \cite{trimmer1989microrobot} to demonstrate remarkable capabilities, including high-speed running \cite{haldane2015running}, jumping \cite{haldane2018robotic}, and climbing \cite{birkmeyer2012dynamic}. However, these results have largely been achieved using simplified models \cite{hoffman2012turning} and time-consuming experimentation \cite{goldberg2017high2, goldberg2017gait} due to the mechanical complexity of the robots and challenges associated with modeling legged systems. 

The ability to effectively model and control legged microrobots could augment these experimental approaches and greatly benefit work in this area. We evaluate a model-based optimization method for designing agile locomotion behaviors involving contact that avoids the need for exhaustive experimentation. These model-based tools are increasingly important as microrobots move closer to envisioned applications, including inspection in confined  environments \cite{JayaramE950}, search and rescue, and environmental monitoring. This work is also broadly applicable to other systems that interact with the environment through contact, including manipulators and larger legged robots. 

\begin{figure}
	\begin{center}
		\includegraphics[width=\columnwidth]{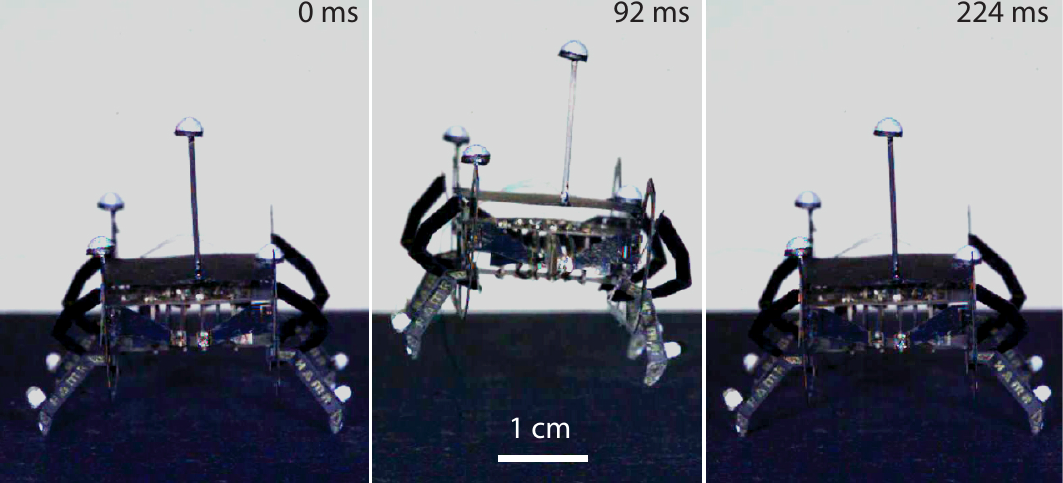}
		\vspace{-0.7cm}
		\caption{Still frames from the supplementary video (\url{https://youtu.be/x_wbRQpKukg}) of HAMR-VI executing a vertical jump.} 
		\label{fig:introfig}
		\vspace{-0.75cm}
	\end{center}
\end{figure}

\subsection{Related Work}
A variety of sophisticated model-based methods have been developed to design trajectories for legged robots, including hybrid \cite{Farshidian16,Mombaur09,Posa16,reher2016algorithmic,Remy11} and contact-implicit \cite{manchester2017variational, Mordatch12, Posa14, Tassa12} trajectory optimization methods. Contact-implicit methods have the benefit of generating contact sequences as part of the optimization, eliminating the need for \emph{a-priori} contact mode scheduling. This enables planning for a variety of periodic and aperiodic behaviors. Most methods, however, are limited to first-order integration accuracy, which creates a linear trade-off between the size of the trajectory optimization problem and the resulting trajectory accuracy. For high-dimensional robots with multiple contacts, the generation of accurate motion plans can lead to nonlinear programs with  tens of thousands of variables that push the limits of modern solvers. In practice, coarse time discretizations are used to reduce program size, resulting in inaccurate whole-body locomotion plans that are difficult or impossible to realize on a physical robot. 

In spite of this drawback, contact-implicit methods are often used to plan for reduced systems, such as zero-moment point or centroidal dynamics models. These low-dimensional trajectories are then stabilized using an inverse dynamics controller that resolve whole-body motions online \cite{righetti2011inverse}. Such methods have achieved success on the humanoid Atlas robot \cite{kuindersma2016optimization}, and the quadrupedal HyQ robot \cite{winkler2015planning, winkler2017fast}. Another approach, also implemented on the HyQ robot, uses relaxed or spring-damper contact models specifically calibrated for a given system \cite{Neunert17}. Despite some successes, whole-body motion plans produced using general physics-based models have yet to be realized on a physical robot. 

Recently, \citet{manchester2017variational} developed a variational contact-implicit trajectory optimization scheme that combines ideas from discrete variational mechanics with the complementarity formulation of rigid-body contact to achieve higher-order integration accuracy. Simulations from that work demonstrate that this approach results in more accurate whole-body locomotion plans that can be tracked with simple controllers. In this paper, we extend and apply these methods to plan and track locomotion trajectories on a physical microrobot. 

\subsection{Contributions}

Our primary contribution is the development and evaluation of a framework for modeling, planning, and controlling dynamic behaviors for legged microrobots. We develop a full three-dimensional dynamic model of the Harvard Ambulatory MicroRobot (HAMR-VI), a quadrupedal microrobot with eight control inputs, 76 states, and 24 kinematic position constraints \cite{doshi2015model}. We also adapt a state-of-the-art variational contact-implicit trajectory optimization algorithm to generate physically accurate locomotion plans. Finally, we develop a low-latency estimator and joint-space controller that allow this microrobot to track the generated plans. 

Our methods are used to generate nine periodic gaits at three frequencies (\SI{2}{\hertz}, \SI{10}{\hertz}, and \SI{30}{\hertz}) on three different surfaces (sandpaper, card-stock, and Teflon). Optimized gaits are shown to move 17\% faster per-cycle on average than manually tuned gaits at the same operating conditions. We also execute a gait with an average velocity of \SI{10.87}{\mm}/cycle, the fastest recorded gait for this platform. Finally, we demonstrate the first controlled vertical jump of \SI{9.96}{\milli\meter}, which is 78\% of the microrobot's center-of-mass (COM) height.

\subsection{Paper Organization}

The remainder of this paper is organized as follows: we present an overview of the platform and a dynamic model of the system in Sec. \ref{sec:model}. The trajectory optimization problem for periodic and aperiodic behaviors is formalized in Sec. \ref{sec:trajopt}. In Sec. \ref{sec:locomotion_exp}, we describe the hardware and software used for the locomotion experiments, and we present and discuss the results of these experiments in Sec. \ref{sec:results} and Sec. \ref{sec:discussion}, respectively. Finally, we draw conclusions and present directions for future research in Sec. \ref{sec:conclusions}. 

\section{Dynamic Model}
\label{sec:model}

\begin{figure}
	\begin{center}
		\includegraphics[width=\columnwidth]{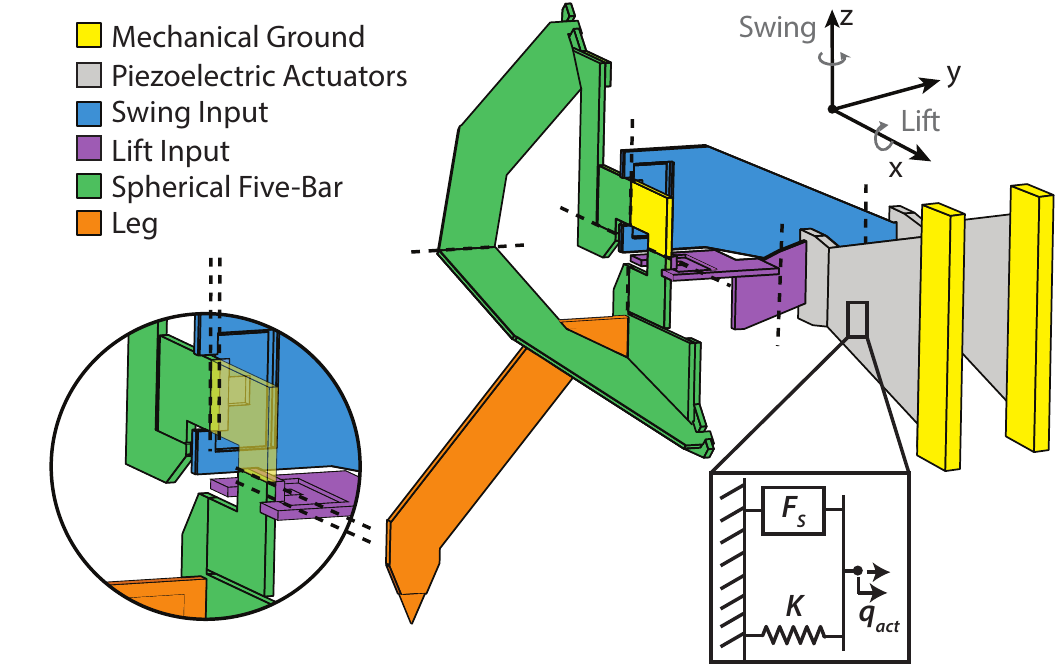}
		\vspace{-0.65cm}
		\caption{Perspective view of the rear-right 2-DOF spherical five-bar (SFB) transmission with components, flexural joints, and body-fixed axes labeled. Circular and rectangular insets depict a detailed view of the SFB's center of rotation, and a simplified mechanical model of the piezoelectric bending actuators, respectively. } 
		\label{fig:modelfig}
		\vspace{-0.75cm}
	\end{center}
\end{figure}

\subsection{Platform Overview}
\label{sec:overview}

HAMR-VI \ (Fig. \ref{fig:introfig}) is a \SI{4.51}{\centi\meter} long, \SI{1.5}{\gram} quadrupedal microrobot with eight independently actuated DOFs. Each leg has two DOFs that are driven by optimal energy density piezoelectric bending actuators (henceforth actuators) \cite{jafferis2015design}. These actuators are controlled with AC voltage signals using a simultaneous drive configuration described by \citet{karpelson2012driving}. A spherical-five-bar (SFB) transmission (Fig. \ref{fig:modelfig}) connects the two actuators to a single leg in a nominally decoupled manner: the swing actuator controls leg-$x$ motion, and the lift actuator controls the leg-$z$ motion. Each SFB transmission has 11 carbon fiber linkages (QA-112, Tohotenax) connected by nine compliant polyimide flexures (Kapton, Dupont) in three parallel kinematic chains.
	

\subsection{Robot Model}
\label{sec:robot}

 The dynamics of the SFB transmissions are assumed to follow the pseudo-rigid body approximation \cite{howell2001compliant}, with the flexures and carbon fiber linkages modeled as pin joints and rigid bodies, respectively. Each flexure is assumed to deflect only in pure bending with its mechanical properties described by a torsional spring and damper that are sized according to the procedure described by \citet{doshi2015model}. Given these assumptions, each SFB transmission has two inputs (forces generated by the actuators), and eight generalized coordinates. These include two independent coordinates (actuator tip deflections) and six dependent coordinates (a subset of flexure joint angles) that represent the kinematics of the three parallel chains. Thus, a complete model of the robot has eight inputs, 38 generalized coordinates (76 states), and 24 position constraints. 
 
A three-dimensional computer-aided-design model is developed using SolidWorks (Dassault Syst\`{e}mes) to capture the kinematics and inertial properties of the microrobot. An open-source SolidWorks to Universal-Robot-Description-Format (URDF) exporter is used to generate an initial URDF model of the microrobot. Actuator forces, joint limits, kinematic-loop constraints, and the mechanics (stiffness and damping) of the flexural joints are then manually incorporated. Furthermore, units are rescaled from SI (seconds, meters, and kilograms) to milliseconds, millimeters, and grams for improved numerical conditioning. The dynamics and control toolbox Drake \cite{Tedrake16} computes the terms in the Euler-Lagrange equation from this URDF description using the composite rigid body algorithm \cite{featherstone1983calculation},
\begin{align}
\label{eqn:forced_lagrangian}
\begin{split}
\frac{d}{dt} D_2 \mathcal{L}(q, \dot{q})  - D_1 \mathcal{L}(q, \dot{q})\ + 
C(q)^T \lambda & = 
\\ F^\textrm{b} (q, \dot{q}) + F^\textrm{act}(q) & 
\end{split} \\
c(q)& =  0. \label{eqn:loop_const}
\end{align}
Here $\mathcal{L}$ is the microrobot's Lagrangian (including flexural spring energy), $q \in \mathbf{R}^{38}$ is the vector of generalized coordinates, $C(q)^T = (\partial c / \partial q)^T$ is the Jacobian mapping constraint forces, $\lambda$, into generalized coordinates, and the dot superscripts represent time derivatives. $F^{\textrm{b}}$ is the vector of generalized flexural damping forces, and $F^{\textrm{act}}$ is the vector of generalized actuator forces. Equation \eqref{eqn:loop_const} enforces the kinematic-loop constraints $c(q)$. Finally, the \emph{slot derivative} $D_i$ indicates partial differentiation with respect to a function's $i^{\text{th}}$ argument.

Each actuator is modeled as a force source in parallel with a spring (Fig. \ref{fig:modelfig}, inset) to determine $F^{\textrm{act}}$. The contribution of the effective mass and damping of the actuator is negligible. An affine approximation of the force source model, and a constant approximation of the spring model developed by \citet{jafferis2015design} are used for simplicity. These approximations have been experimentally verified for the range of expected operating voltages ($\sim$\SI{100}{}-\SI{200}{\volt}) \cite{doshi2015model}. The generalized actuator force can then be written as,
\begin{align}
\label{eqn:act}
F^{\textrm{act}}(u, q) = B(q)^T (F_{\textrm{s}}(u, q) - K q),
\end{align}
where $F_s$ is the piezoelectric force, $K$ is the spring stiffness, $u$ is the AC drive voltage, and  $B^T$ is the Jacobian mapping actuator forces into generalized coordinates. Substituting \eqref{eqn:act} into \eqref{eqn:forced_lagrangian} and \eqref{eqn:loop_const} gives a complete set of differential-algebraic equations that capture the dynamics of the microrobot:
 \begin{align}
\label{eq:complete_forced_lagrangian}
\centering
\begin{split}
\frac{d}{dt} D_2 \mathcal{L}(q, \dot{q}) - D_1 \mathcal{L}(q, \dot{q}) + 
C(q)^T \lambda &= \\
F^\textrm{b}(q, \dot{q}) + B(q)^T (F_{\textrm{s}}(u, q) - Kq)\\
c(q) &= 0.
 \end{split}
\end{align}



\subsection{Surface Characterization}
\begin{figure}
	\begin{center}
		\includegraphics[width=\columnwidth]{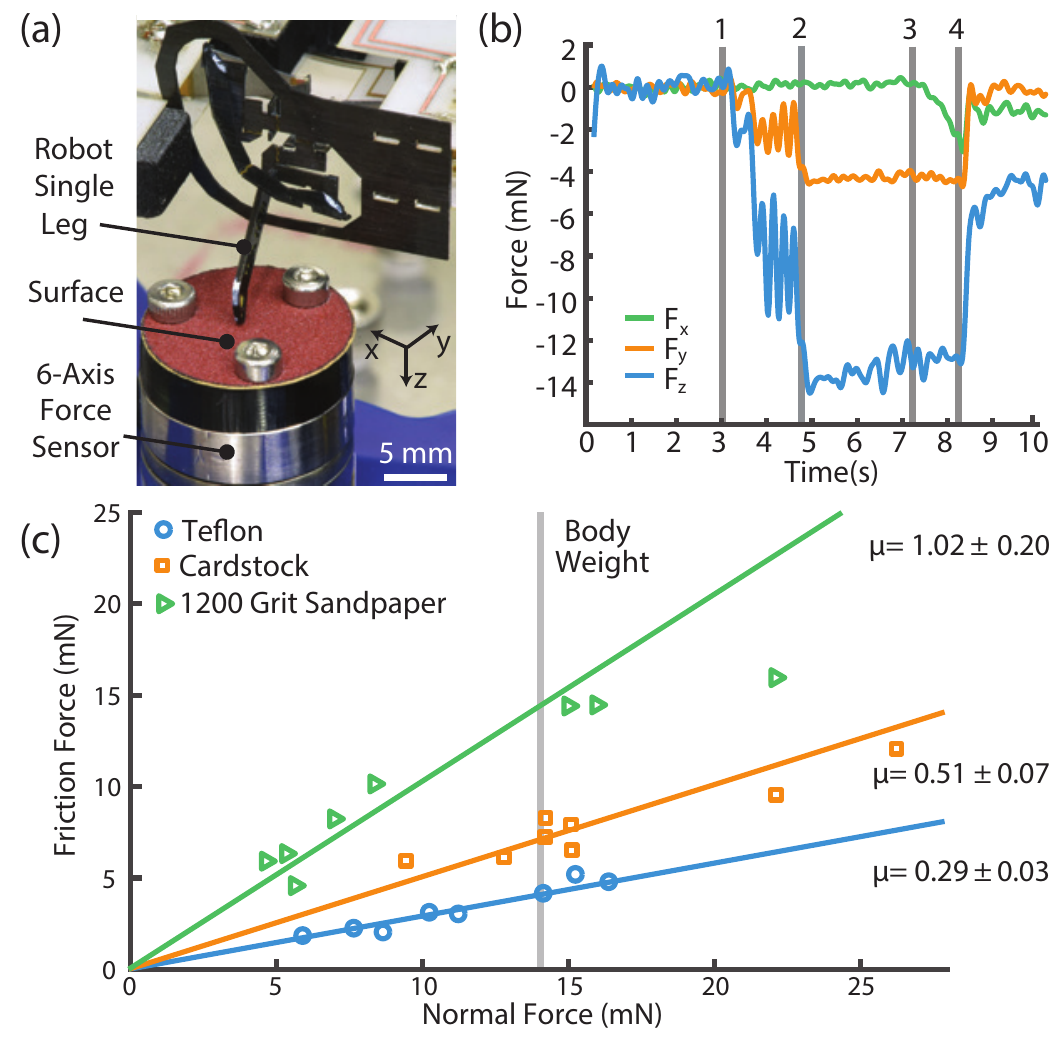}
		\vspace{-0.7cm}
		\caption{(a) Labeled image of the friction-measurement experimental setup. (b) Representative force data: F$_\text{z}$ is the normal force, and  F$_\text{x}$ and F$_\text{y}$ are the tangential forces.  (c) Raw data (n=8) and best-fit lines corresponding to the estimated coefficients of static friction between the microrobot's foot and the Teflon, cardstock, and 1200 grit sandpaper surfaces.}
		\label{fig:surfaceexp}
		\vspace{-0.75cm}
	\end{center}
\end{figure}

\label{sec:surface}
In addition to modeling the robot's dynamics, the coefficients of static friction ($\mu$) between microrobot's feet and three surfaces, PTFE (Teflon), card-stock, and 1200 grit sandpaper, are measured. Experiments are conducted using a single leg (Fig. \ref{fig:surfaceexp}a) to closely replicate conditions during locomotion. Each surface is placed on an acrylic mounting plate and fastened to a six-axis force sensor (ATI Nano17 Titanium). The single leg is mounted on two micro-positioning stages and centered above the force sensor. Eight trials are run on each surface, and force data is recorded at \SI{100}{\hertz} using  MATLAB's xPC environment (MathWorks, MATLAB R2015a). Force traces for a representative trial are shown in Fig. \ref{fig:surfaceexp}b. The leg is manually lowered to pre-load force sensor to 35\%-200\% of the microrobot's body weight (between 1 and 2 in Fig. \ref{fig:surfaceexp}b). The swing DOF is then actuated, generating a force in the $x-y$ plane (3 in Fig. \ref{fig:surfaceexp}b) until the leg begins to slip (4 in Fig. \ref{fig:surfaceexp}b). Force data is filtered using an acasual low-pass Butterworth filter with a cutoff frequency of \SI{10}{Hz}. The normal, $F_n$, and static frictional, $F_f$, forces are computed as:
\begin{align}
\label{eqn:expfric}
 F_n = \Delta F_z \\
 F_f = \sqrt{\Delta F_x^2 + \Delta F_y^2}, 
\end{align}
where $\Delta F_x$, $\Delta F_y$, and $\Delta F_z$ are the net forces between stages 1 and 4 in Fig. \ref{fig:surfaceexp}b in the $x$, $y$, and $z$ directions, respectively. The friction force increases linearly with the normal force as anticipated (Fig. \ref{fig:surfaceexp}c). The mean and standard deviation for coefficients of friction averaged over the eight trials for Teflon, card-stock, and 1200 grit sandpaper are: 0.29 $\pm$ 0.03, 0.51$\pm$ 0.07, and 1.02 $\pm$ 0.20, respectively. Lines corresponding to these average friction coefficients are shown in Fig. \ref{fig:surfaceexp}c. 

\subsection{Contact Model}
\label{sec:contact}
The robot's four feet are modeled as point contacts. The contact forces are decomposed into directions normal and tangential to the running substrate. In the normal direction, collisions must obey a non-penetration constraint: 
 \begin{align}
 \label{eqn:normal}
 \phi(q) \geq 0,
 \end{align} 
where $\phi (q)$ is a function that returns the signed distance between the four feet and the running substrate. Tangential forces are modeled using Coulomb fricton, and they must satisfy the \textit{Maximum Dissipation Principle} \cite{Moreau73}. This states that the friction force instantaneously maximizes the dissipation of kinetic energy, and is  a generalization of the 2D concept of friction opposing the direction of motion. Mathematically, this can be formulated as the following optimization problem:
 \begin{align}
 \label{eqn:tangential}
 \begin{split}
 \begin{aligned}
 &\underset{b}{\text{minimize}} 
 & & \dot{q}^T D(q)^T b \\
 &\text{subject to} 
 & &||b|| \leq \mu \gamma,
 \end{aligned}
 \end{split}
 \end{align} 
 where $b$ is the friction force, $\gamma$ is the normal force, $\mu$ is the coefficient of friction, and $D^T$ is the Jacobian mapping tangential contact forces into generalized coordinates. The norm constraint on $b$ ensures that the friction force lies within the Coulomb friction cone.
 
Instead of using the linear complimentarity formulation described by \citet{Stewart96} to combine the normal \eqref{eqn:normal} and tangential \eqref{eqn:tangential} contact constraints with the microrobot dynamics \eqref{eq:complete_forced_lagrangian}, we adapt the \emph{variational} framework recently developed by \citet{manchester2017variational}. The variational framework requires approximating the Lagrange-D'Alembert principle with a quadrature rule before taking variational derivatives \cite{Marsden99}. The order of the resulting discrete equations-of-motion depends on the choice on quadrature rule, and we use the midpoint rule for second-order accuracy. 

The normal contact constraints \eqref{eqn:normal} and the microrobot's kinematic constraints \eqref{eqn:loop_const} are added to the discrete Lagrange-D'Alembert principle with the appropriate Lagrange multipliers, $\gamma$ (the normal force) and $\lambda$ (the constraint force), respectively. Taking variations with respect to the generalized coordinates $q_k$ then leads to the discrete Euler-Lagrange equations for our system: 
\begin{equation} \label{eq:KKT1}
\begin{split} 
&D_2 \mathcal{L}_d(h, q_{k-1}, q_k) + D_1 \mathcal{L}_d(h, q_k, q_{k+1})  \\ 
&+ \frac{1}{2} F^\textrm{ext}_d(h, q_{k-1}, q_k, u_{k-1}) + \frac{1}{2} F_d^\textrm{ext}(h, q_k, q_{k+1}, u_{k}) \\ 
& + \frac{1}{2} (C(h, q_{k-1}, q_k) + C(h, q_k, q_{k+1}))^T \lambda_k \\ 
&+ N(q_{k+1})^T \gamma_k = 0. 
\end{split}
\end{equation}
Here $\mathcal{L}_d$ is the discrete Lagrangian, defined as the midpoint approximation of the integral of the continuous Lagrangian over a single time step. $F^\textrm{ext}_d$ is an analogous discretization of the generalized external force 
\eqref{eqn:act}. Note that, in general, $\mathcal{L}_d$ and $ F^{\textrm{ext}}_d$ depend on the choice of quadrature rule, and expressions for both are derived for a general system using the midpoint rule in \cite{manchester2017variational}. Finally, $N(q)^T = (\partial \phi / \partial q)^T$ is the Jacobian mapping normal contact forces into generalized coordinates. 

In addition to \eqref{eq:KKT1}, solutions must satisfy the following constraints, known as Karush-Kuhn-Tucker (KKT) conditions \cite{Boyd04}:
\begin{align} 
\label{eqn:disc_loop}
c_d(q_k, q_{k+1}) = 0 & \\ 
\begin{split} \label{eq:KKT2}
\gamma_k \geq 0 &\\ 
\phi(q_{k+1}) \geq 0 &\\
\gamma_k^T \phi(q_{k+1}) = 0, &
\end{split}
\end{align}
where \eqref{eqn:disc_loop} ensures that kinematic constraints are enforced, and $c_d$ is the midpoint approximation of $c$. The three conditions in \eqref{eq:KKT2}, collectively known as a \emph{complementarity constraint}, prevent interpenetration and ensure that contact forces act only when bodies are in contact to push them apart. Such constraints are commonly denoted using the following shorthand notation:
\begin{equation} \label{eq:perp}
0 \leq \gamma_k \perp \phi(q_{k+1}) \geq 0.
\end{equation}

The KKT optimality conditions for an approximation of the \textit{Maximum Dissipation Principle} \eqref{eqn:tangential} are derived and discretized using the midpoint rule in a similar manner. This results in three additional constraints: one equality constraint and two complementarity constraints, 
\begin{align} \label{frictionKKT}
\begin{split}
g_1(h, q_k, q_{k+1}, \psi_k, \eta_k) = 0 \\
0 \leq \psi_k \perp g_2(\gamma_k, \beta_k) \geq 0 \\
0 \leq \beta_k \perp \eta_k \geq 0, 
\end{split}
\end{align} 
where $\psi$ and $\eta$ are Lagrange multipliers and the exact forms of $g_1$ and $g_2$ are  derived in \cite{manchester2017variational}. The tangential force and corresponding constraints are added to \eqref{eq:KKT1} and \eqref{eq:KKT2} to complete the dynamics model,
\begin{align} \label{eq:finalmodel}
\begin{split} 
& r(h, q_{k-1}, q_k, q_{k+1}, u_{k-1}, u_k, \lambda_k, \gamma_k) +  P(q_{k+1})^T \beta_k    = 0 \\
& g_1(h, q_k, q_{k+1}, \psi_k, \eta_k) = 0 \\
& c(q_k, q_{k+1}) = 0 \\
& 0 \leq \gamma_k \perp \phi(q_{k+1}) \geq 0. \\
& 0 \leq \psi_k \perp g_2(\gamma_k, \beta_k) \geq 0 \\
& 0 \leq \beta_k \perp \eta_k \geq 0.
\end{split}
\end{align}
Here, $r$ is the LHS of \eqref{eq:KKT1}, and $\beta$ and $P$ are defined in \cite{manchester2017variational} based on b and $D$. Given $q_{k-1}$,  $q_k$, $u_{k-1}$, and $u_k$, \eqref{eq:finalmodel} can be solved to find $\gamma_k$, $\lambda_k$, $\beta_k$, $\psi_k$, $\eta_k$, and $q_{k+1}$.

\section{Trajectory Optimization}
\label{sec:trajopt}

\begin{figure}
	\begin{center}
		\includegraphics[width=\columnwidth]{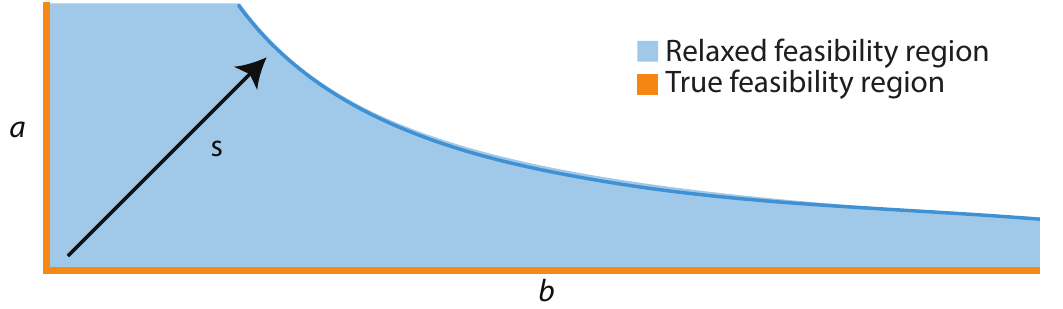}
		\vspace{-0.75cm}
		\caption{Relaxed (blue) and true (orange) feasibility regions for a complimentarity constraint of the form $0 \leq a \perp b \geq 0$. Note that we recover the true feasible region when $s=0$. } 
		\label{fig:comp_slack}
		\vspace{-0.75cm}
	\end{center}
\end{figure}

The dynamics expressed in \eqref{eq:finalmodel} are used as constraints in a direct trajectory optimization scheme. The trajectory optimization problem is posed as a standard nonlinear program (NLP) and solved using the constrained optimization solver SNOPT. To ease the numerical difficulties associated with complementarity constraints, we apply a smoothing scheme similar to that used in \cite{Fletcher04}.  For a complimentarity constraint of the form $\leq a \perp b \geq 0$, this smoothing scheme replaces the equality constraint $a^T b = 0$ with the inequality constraint $a^T b - s \leq 0$. Here, $s$ is a non-negative slack variable that alters the feasible region as shown in Fig. \ref{fig:comp_slack}. 
We use a single slack variable, $s_k$, to smooth all complimentarity constraints at each knot point. To encourage convergence of solutions towards satisfaction of the true complementarity constraints, we augment the cost function with a term that penalizes $s_k$.
The complete formulation of the trajectory optimization problem is stated as the following NLP:
\begin{equation} \label{eq:NLP}
\begin{aligned}
& \underset{h,\, \mathcal{Q},\, \mathcal{U},\, \mathcal{C}}{\text{minimize}}  & J(h, \mathcal{Q}, \mathcal{U}) + \alpha \sum_{k=1}^{N-1} s_k \,\,\,\,\,\,\,\,\,\,\,\, \\
& \text{subject to} & f(h, \chi) = 0& \\
& & g(q_{k+1}, \lambda_k, \beta_k, \psi_k, \eta_k, s_k) \geq 0& \\
& & u_{\mathrm{min}} \leq u_k \leq u_{\mathrm{max}}& \\
& & q_{\mathrm{min}} \leq q \leq q_{\mathrm{max}}& ,
\end{aligned}
\end{equation}
where $J$ is a cost function, $\alpha$ is a positive scalar weighting parameter, $f$ and $g$ are the equality and inequality constraints in the relaxed version of \eqref{eq:finalmodel} found in \cite{manchester2017variational}, and $\chi =\{ q_{k-1}, q_k, q_{k+1}, u_{k-1}, u_k, \lambda_k, \gamma_k, \beta_k, \psi_k, \eta_k \}$. Furthermore, $\mathcal{Q}$ is the set of robot configurations at the knot points, $q_k$, $\mathcal{U}$ is the set of control voltages, $u_k$, and $\mathcal{C}$ is the set of all constraint-related variables, $\lambda_k$, $\gamma_k$, $\beta_k$, $\psi_k$, $\eta_k$, and $s_k$. The input voltages, $u_k$, are bounded by $u_{\mathrm{min}}=$ \SI{0}{\volt} and $u_{\mathrm{max}}=$ \SI{225}{\volt} to increase actuator lifetime \cite{jafferis2015design}. The flexure joint angles, a subset of $q_k$, are bounded between $q_{\mathrm{min}}= -\pi/4$ and $q_{\mathrm{max}}= \pi/4$, which are conservative estimates of the maximum mechanical bend angles. The penalty on the slack variables in the cost function of \eqref{eq:NLP} is an ``exact penalty'' that has theoretical convergence guarantees with finite values of $\alpha$~\cite{Anitescu05}. In practice, we find good convergence with values of $\alpha$ on the order of $10^2$.

\subsection{Gait Optimization}
\label{sec: periodic}

We search for gaits near stride frequencies of \SI{2}{\hertz}, \SI{10}{\hertz}, and \SI{30}{\hertz} on three different surfaces: Teflon, card-stock, and 1200 grit sandpaper. The selected frequencies represent different operational regimes for the microrobot as discussed by \citet{goldberg2017gait}: quasi-static (\SI{2}{\hertz}), near the $z$-natural frequency (\SI{10}{\hertz}), and near the roll natural frequency (\SI{30}{\hertz}). In addition, the \SI{2}{\hertz} gaits represent a long time horizon for this microrobot as the body natural frequencies are between 10--\SI{30}{\hertz}. Finally, these nine gaits cover a wide-range of (ground) contact conditions ($\mu\in [0.29,1.02]$), showcasing the versatility of our approach. 

The NLP presented in \eqref{eq:NLP} is modified to search for periodic state and input trajectories by enforcing periodicity constraints on all position and velocity decision variables except the $x$-position of the floating base. The algorithm minimizes the following cost function that encourages the robot to achieve its maximum theoretical stride length: 
\begin{equation}
\label{eqn:periodic_cost}
\begin{split}
J = (x_N - x_g)^T& Q(x_N - x_g) \\ 
& + \sum_{i=2}^{N-1} \frac{c_1}{2} \Delta \dot{q}_i^T\Delta \dot{q}_i +  \frac{1}{2}\Delta u_i^T \Delta u_i,
\end{split}
\end{equation} 
where $x_g$ is a goal state, $Q$ is a diagonal matrix with $Q_{11} \in [10,50]$ and the remaining diagonal entries equal to one, $c_1\in [10,50]$ scales the velocity difference penalty, and $\Delta \dot{q}_i = \dot{q}_{i} - \dot{q}_{i-1}$ and $\Delta u_i = u_{i} - u_{i-1}$ are the difference between subsequent generalized velocities and control voltages, respectively. To reduce sensitivity to local minima, the optimization is initialized with a heuristic trot gait that achieves roughly periodic locomotion assuming cardstock friction. The goal state is then defined as $x_g = [10, x_p]^T$, where the first entry corresponds to translating the body forward slightly less than twice the maximum swing displacement in a gait cycle. The periodic subset of the goal state, $x_p \in \mathbb{R}^{75}$, is set to the periodic subset of the final state on the initial trajectory. The difference penalties are applied to discourage chatter in the state and inputs trajectory.

\subsection{Aperiodic Behaviors}
\label{sec: aperiodic}

We also used this variational trajectory optimization method to find state and input trajectories for a vertical jump. The following cost function, which encourages the microrobot to jump to a specific height, is minimized: 
\begin{equation}
\label{eqn:jump_cost}
\begin{split}
J = (x_N - x_g)^T& Q(x_N - x_g) + \sum_{i=1}^{N-1} \frac{1}{2} u_i^T R u_i.
\end{split}
\end{equation} 
Here $x_g = [0_{2\times1},  24, 0_{73\times1}]^T$ is a goal state that specifies the desired apex height of the jump (a little less than twice the COM height) with no body rotation or horizontal motion. The quadratic input cost penalizes swing actuator voltages as fore/aft forces do not contribute significantly to a vertical jump. To improve convergence time and avoid poor local optima, the optimization is initialized with a heuristically designed vertical jump trajectory.  

\section{Locomotion Experiments}
\label{sec:locomotion_exp}

\begin{figure*}
	\begin{center}
		\includegraphics[width=7in]{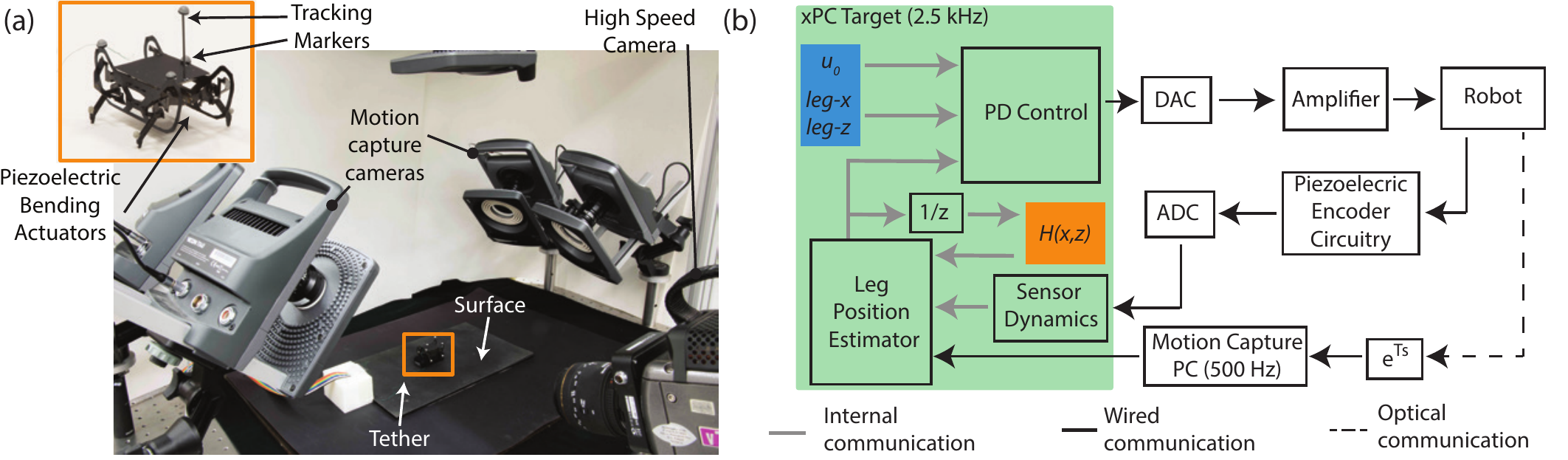}
		\vspace{-0.25cm}
		\caption{(a) Perspective image of locomotion arena with components labeled. (b) Communication and control block diagram for periodic and dynamic locomotion experiments. The leg position estimator and PD controller run on the xPC target (green) at \SI{2.5}{\kilo\hertz}. Feed-forward input voltages and desired leg positions (blue) are pre-computed from the optimized trajectory, and $H(x,z)$ (orange) is a lookup table whose construction is described in Sec. \ref{sec:cont_and_est}.} 
		\label{fig:exp_setup}
		\vspace{-0.75cm}
	\end{center}
\end{figure*}

The microrobot's performance for trajectories found in Sec. \ref{sec:trajopt} is evaluated in a controlled \SI{20}{\centi\meter} $\times$ \SI{20}{\centi\meter} locomotion arena (Fig \ref{fig:exp_setup}a). A proportional-derivative (PD) controller is implemented to track the desired positions of the microrobot's four legs in the body-fixed frame, and an estimator is developed to provide low latency estimates of the leg positions (Fig. \ref{fig:exp_setup}b).


\subsection{Locomotion Arena}
\label{sec:arena}

Input signals are generated at \SI{2.5}{\kilo\hertz} using a MATLAB xPC environment (MathWorks, MATLAB R2015a), and are supplied to the microrobot through a ten-wire tether. Five motion capture cameras (Vicon T040) track the position and orientation of the robot body and the position of the feet at \SI{500}{\hertz} with a latency of \SI{11}{\milli\second}. In addition, eight piezoelectric encoders (described below) provide low-latency estimates of actuator tip velocities at \SI{2.5}{\kilo\hertz} \cite{jayaram2018concomitant}. Finally, a high speed camera (Phantom v7.3) orthogonal to the sagittal plane records the motion of the robot at \SI{500}{\hertz}. 

\subsection{Piezoelectric Encoder Dynamics}
\label{sec:piezo_dyn}

\begin{figure}
	\begin{center}
		\includegraphics[width=\columnwidth]{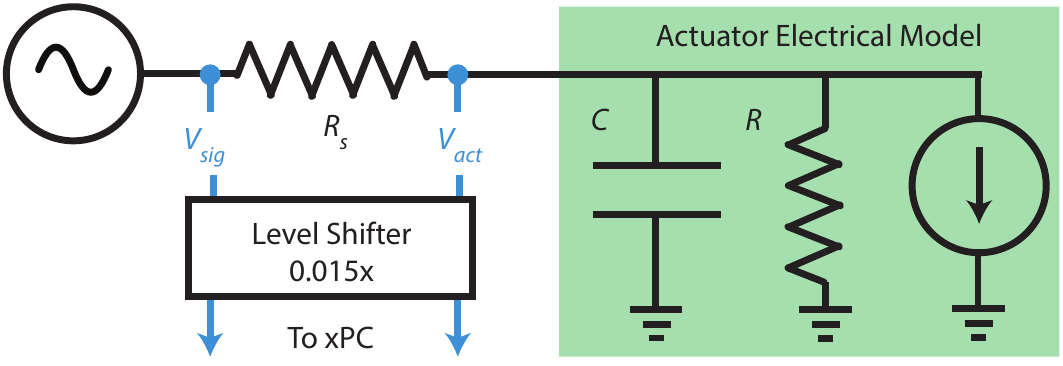}
		\vspace{-0.5cm}
		\caption{Schematic of lumped parameter electrical model of a single actuator (green) and associated piezoeletric encoder  measurement circuit \cite{jayaram2018concomitant}.} 
		\label{fig:piezo_encoder}
		\vspace{-0.75cm}
	\end{center}
\end{figure}

Each piezoelectric encoder (Fig. \ref{fig:piezo_encoder}) provides an estimate of the corresponding actuator's tip velocity by computing the ``mechanical" current (proportional to tip velocity \cite{dosch1992self}) produced in that actuator. This current, $i_\text{m}$, can be computed by applying Kirchoff's law \eqref{eqn:sensor_dyn} to the measurement circuit in series with a lumped-parameter electrical model of an actuator:   
\begin{align}
\label{eqn:sensor_dyn}
i_\text{m} = & \frac{u_\text{sig} - u_\text{act}}{R_s} -  \zeta C \dot{u}_\text{act} - \frac{u_\text{act}}{R}.
\end{align}
The first term on the RHS is the total current drawn by an actuator, which is computed on the xPC target from measurements of the voltages before, $u_\text{sig}$, and after, $u_\text{act}$, a shunt resistor, $R_{s}$. The actuator is modeled as a capacitor, $C$, resistor, $R$, and current source, $i_\text{m}$, in parallel. The voltage and frequency dependent values of $R$ and $C$, and the value of the blocking factor $\zeta$ have been computed by \citet{jayaram2018concomitant}.

\subsection{Controller and Estimator Design}
\label{sec:cont_and_est}

\begin{figure*}
	\begin{center}
		\includegraphics[width=7in]{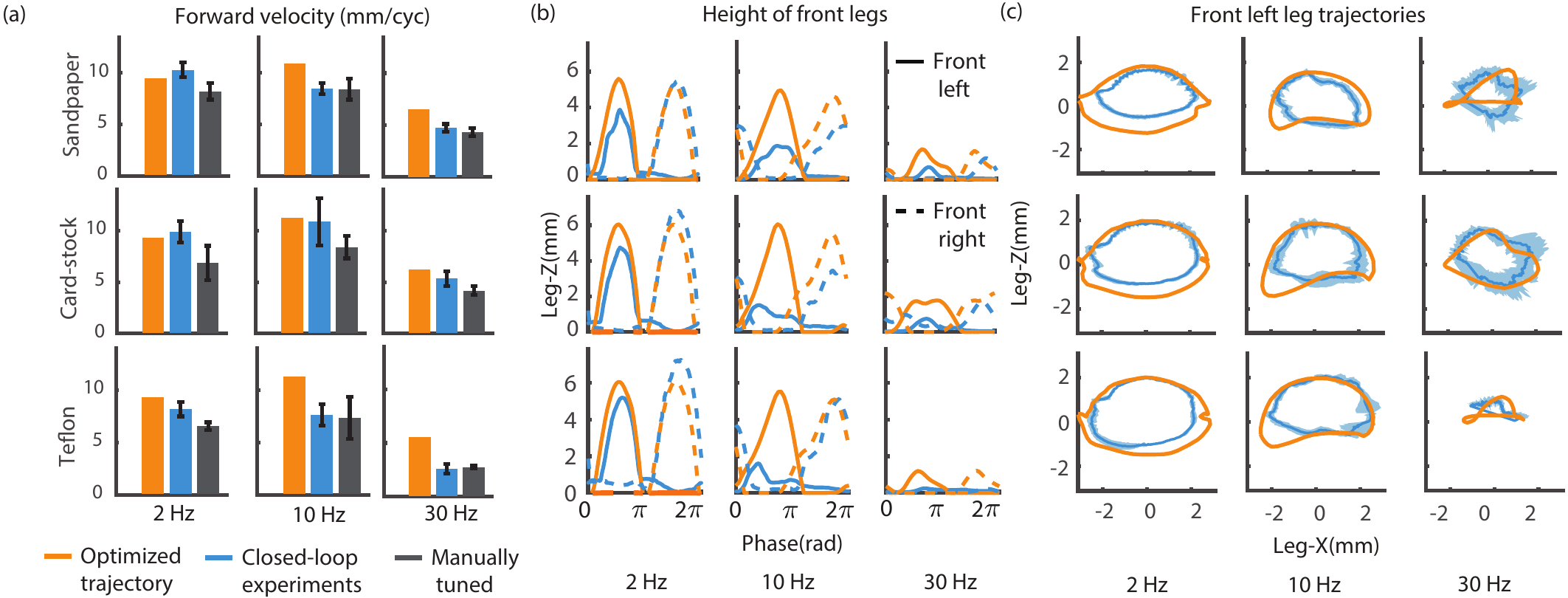}
		\vspace{-0.25cm}
		\caption{(a) Mean per-cycle forward velocity for the optimized (orange), closed-loop experimental (blue), and manually tuned (gray) trajectories. Error bars represent one standard deviation (n=15). (b) Optimized (orange) and mean closed-loop experimental (blue) leg height for the front left (solid) and front right (dashed) leg. (c) Optimized (orange) and mean closed-loop experimental (blue) periodic foot trajectories for the front left leg. The blue shaded region represents one standard deviation (n=15).}  
		\label{fig:results1}
		\vspace{-0.75cm}
	\end{center}
\end{figure*}

Since the motion capture measurement latency is a significant percentage of the gait period ($\sim$30\% at \SI{30}{\hertz}), a filter is developed to provide low-latency leg position estimates in the body-fixed frame by using the current measurements from the piezoelectric encoders. These currents are scaled by an empirically determined proportionality constant, $X$, to estimate actuator velocities. For a particular leg, the SFB transmission kinematics are used to define a transformation, $H \in \mathbf{R}^{3\times2}$, from lift and swing actuator velocities to Cartesian leg velocities in the body-fixed frame. This map is used to estimate Cartesian leg velocity in the body-fixed frame, which is integrated over the duration of the latency to achieve low-latency leg position measurements. 

The map \eqref{eq:trans_map} is indexed by achievable leg-$x$ and leg-$z$ positions (instead of lift and swing actuator positions) as they are directly estimated,  
\begin{align}
\label{eq:trans_map}
H(x,z) = \frac{\partial f}{\partial q_i}  - \frac{\partial f}{\partial q_d} \bigg[\frac{\partial c}{\partial q_d}\bigg]^{-1}  \frac{\partial c}{\partial q_i}. 
\end{align}
Here, $q_i$ are the independent generalized coordinates (actuator positions), $q_d$ are the dependent generalized coordinates (flexure joint angles), $c$ is the vector of kinematic loop constraints, and $f$ is the kinematic mapping from lift and swing actuator positions to Cartesian leg position. We solve an inverse kinematics problem, posed as a NLP, to find values of $q_i$ and $q_d$ that result in desired leg-$x$ and leg-$z$ values. Twenty-one samples are used in each direction, and $H(x,z)$ is defined for each leg as a $3 \times 2$ look-up table stored on the xPC target.  

This look-up table is then used in the following estimator: 
\begin{align}
\label{eqn:estimator1}
\hat{q}[k]_\text{leg} = & q[k-\epsilon]_\text{leg} + T_s \sum_{\kappa = k-\epsilon}^k \frac{1}{2}( \dot{\hat{q}}[\kappa]_\text{leg} -  \dot{\hat{q}}[\kappa-1]_\text{leg}) \\
\label{eqn:estimator2}
 \dot{\hat{q}}[k]_\text{leg} =&  H(x[k-1], z[k-1])  \begin{bmatrix} X^s & 0 \\0 & X^l \end{bmatrix}  \begin{pmatrix} i_m^s[k] \\i_m^l[k]\end{pmatrix}.  
\end{align}
In \eqref{eqn:estimator1}, $\hat{q}[k]_\text{leg}$ is the estimated leg position,  $q[k-\epsilon]_\text{leg}$ is leg position measured by the motion capture system with $\epsilon$ latency, $T_s$ is the sample rate of the xPC Target, and the summation is a trapezoidal integration of the estimated leg velocity over the duration of the latency. In \eqref{eqn:estimator2}, $H$ is indexed by the previous estimate of the leg-$x$ and leg-$z$ positions, and $i[k]_m^s$ and $i[k]_m^l$ are the mechanical currents for the swing and lift actuators, respectively. Finally,  $X^s$ ($\sim$140) and $X^l$ ($\sim$120) are experimentally determined proportionality constants for the swing and lift actuators, respectively,

These estimated leg positions are used in a simple controller since the swing and lift DOFs are nominally decoupled at pre-resonant drive frequencies \cite{doshi2017phase}. This feedback controller simply alters the feed-forward lift and swing input voltages based on the following control law: 
\begin{align}
\label{eqn:controller}
u = \begin{bmatrix} K_f^s & 0 \\0 & K_f^l \end{bmatrix} u_0 + \begin{bmatrix} K_p^s & 0 \\0 & K_p^l \end{bmatrix} e +  \begin{bmatrix} K_d^s & 0 \\0 & K_d^l \end{bmatrix} \frac{de}{dt}.
\end{align}
Here $u = [u_s, u_l]'$  are the input signals provided to the swing and lift actuators, $ u_0 = [u_s^0, u_l^0]'$ are the feed-forward input voltages determined by the trajectory optimization (Sec. \ref{sec:trajopt}), and $e = [e_x, e_z]'$ is the error between the desired and estimated leg-$x$ and leg-$z$ positions. Note that the desired leg positions are pre-computed from the optimized state trajectory using the model kinematics.  Finally, $K_p$, $K_d$, and $K_f$, are the proportional, derivative and feed-forward control gains, respectively, and the superscripts $s$ and $l$ represent the swing \textit{•}and lift DOF, respectively.

%
%
%

\section{Results}

We demonstrate a physical implementation of the generated trajectories for nine periodic gaits and a dynamic jumping behavior. In addition, we show that our simple joint-space controller maintains gait timing and tracks the desired leg trajectories in the body-fixed frame for a range of frequencies on a wide variety of surfaces. 
 
 \label{sec:results}

\subsection{Gait Optimization}

Each gait is executed for 15 cycles with an initial voltage ramp, and the control gains are manually tuned. The feed-forward vary from $0.9$ to $1.1$, reflecting the utility of the optimized inputs. The mean per-cycle forward velocity for the closed-loop experimental trials (blue) is compared with the optimized trajectory (orange) and a manually tuned gait with the same frequency and average input power (gray) in Fig. \ref{fig:results1}a.

\begin{figure}
	\begin{center}
		\includegraphics[width=\columnwidth]{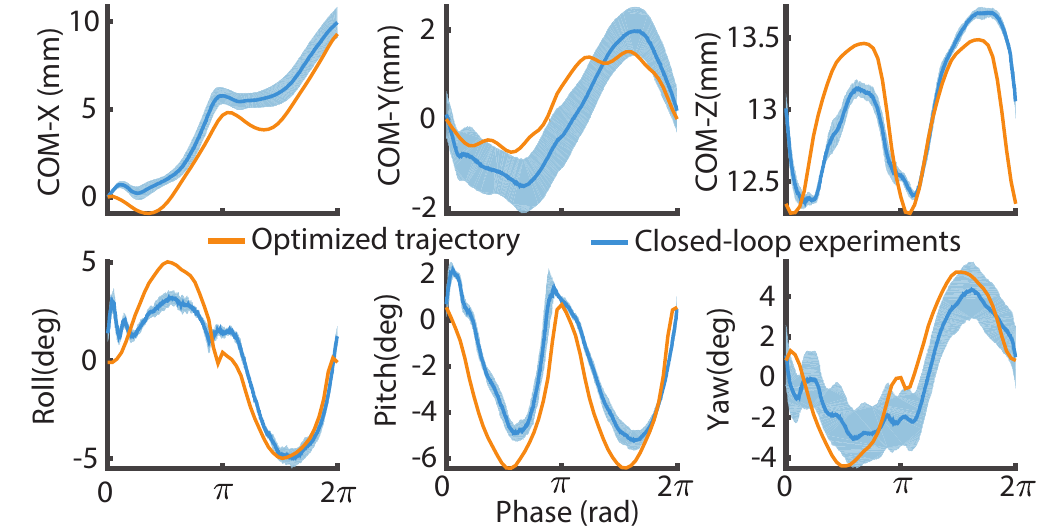}
		\vspace{-0.5cm}
		\caption{ Per-cycle trajectories for the robot body during a \SI{2}{\hertz} gait on card-stock. A best-fit line is subtracted from the $y$ and yaw trajectories. The blue shaded region represents one standard deviation (n=15). }  
		\label{fig:results0}
		\vspace{-0.75cm}
	\end{center}
\end{figure}

The \SI{2}{\hertz} closed-loop experimental trajectories achieve an average velocity of \SI{9.77}{\milli\meter}/cycle, which is within 5\% of the goal speed of \SI{10}{\milli\meter}/cycle. These gaits also perform 26\% better than the manually tuned gaits, which achieve an average velocity of \SI{7.71}{\milli\meter}/cycle. In addition, the planned body pose closely matches that which is executed by the robot (card-stock trial shown in Fig. \ref{fig:results0}), demonstrating that the model captures most of the microrobot's salient dynamic properties. The robot also maintains the desired gait timing (front legs depicted in Fig. \ref{fig:results1}b), and the closed-loop leg trajectories (front left leg depicted in Fig. \ref{fig:results1}c) closely match the optimized trajectories in air ($z > 0$). However, the robot is unable to push as forcefully into the ground as planned, most likely  because of unmodeled serial compliance in the transmissions \cite{ozcan2014powertrain}. Note that the matching of gait timing and acceptable leg-tracking is also consistent for all \SI{10}{\hertz} trajectories and all \SI{30}{\hertz} trajectories, except on Teflon.

At \SI{10}{\hertz}, the closed-loop experimental trajectories achieve an average velocity of \SI{8.98}{\milli\meter}/cycle, which is close to the desired velocity and 10\% faster than the manually tuned gaits. The card-stock gait at this frequency achieves the fastest per-cycle velocity recorded for this robot, at \SI{10.87}{\milli\meter}/cycle. However, the other two gaits perform slightly worse than expected. This is most likely due to discrepancies between the planned and executed floating base trajectories (see supplementary video). Finally, the average velocity for the \SI{30}{\hertz} closed-loop gaits is slower at \SI{4.24}{\milli\meter}/cycle. The closed-loop experiments on sandpaper and card-stock are still 20\% percent faster than the manually tuned gaits and within 20\% of the predicted optimized velocities; however, the robot's performance is poor on Teflon. This frequency (near the roll resonance) is particularly challenging for locomotion using the laterally asymmetric trot gait, and the optimizer has difficulty finding gaits that move at \SI{10}{\milli\meter}/cycle. 
 
\subsection{Aperiodic Behaviors}

\begin{figure}
	\begin{center}
		\includegraphics[width=\columnwidth]{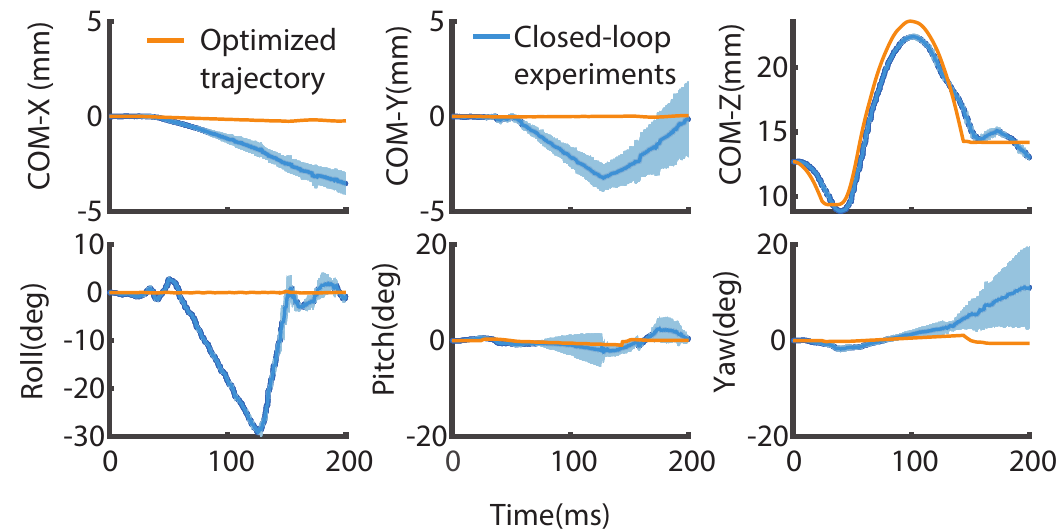}
		\vspace{-0.75cm}
		\caption{ Experimental (blue) and optimized (orange) floating base pose during a vertical jump. The blue shaded region is one standard deviation (n=4).}  
		\label{fig:results2}
		\vspace{-0.75cm}
	\end{center}
\end{figure}

We also use this method to execute a vertical jump on a card-stock surface. The average jump height is $ 9.65 \pm  0.21 $ \SI{}{\milli\meter}, which is $\sim$80\% of the goal height of \SI{12}{\milli\meter} (Fig. \ref{fig:results2}). Manufacturing imperfections lead to asymmetries that cause the robot to roll during the jump, but it stays near the initial position, with final $x$, $y$ and $yaw$ values of $ -3.25 \pm 0.57 $ \SI{}{\milli\meter}, of $ -0.15 \pm 1.98 $ \SI{}{\milli\meter}, and of $ 11.12 \pm 8.56$ \SI{}{\deg}, respectively. 
 
\section{Discussion}
\label{sec:discussion}

\subsection{Performance Improvements}

Our model-based approach yields improvements over previous experimental results collected in \cite{goldberg2017high2, goldberg2017gait}. Specifically, an average velocity of 9.21 $\pm$ \SI{1.31}{\milli\meter}/cycle achieved across the six gaits at \SI{2}{} and \SI{10}{\hertz} is comparable to the highest previously measured experimental velocity of \SI{9.5}{\milli\meter}/cycle achieved using careful tuning on card-stock surface \cite{goldberg2017high2}. Even the three slower \SI{30}{\hertz} gaits move on average 30\% faster than previously recorded trots at similar frequencies on a card-stock surface. Additionally, the robot is able to achieve a new highest velocity of \SI{10.87}{\mm}/cycle, and demonstrate the first controlled vertical jump of \SI{9.96}{\milli\meter} (78\% of COM height). Importantly, these performance improvements were achieved without exhaustive experimentation: tens of experiments were conducted as opposed to hundreds.

\subsection{Quality of Optimized Trajectories}

We evaluated the quality of the periodic trajectories by measuring the normalized average slip, $\bar{s}$ defined as: 
\begin{align}
\bar{s} = \frac{1}{4 \int_{t_0}^{t_f}v_x(t) dt} \sum_{i=1}^4 \int_{\xi}|v^i_x (t)|dt.
\end{align}
Here, $v^i_x$ is the $x$-velocity of the $i$th leg in the world-fixed frame, and $v_x$ is the $x$-velocity of the center of mass in the world-fixed frame as  measured by the motion capture system. The time interval of interest is bounded by $t_0$ and $t_f$, and $\xi$ is the set of times for which  $v^i_x < 0$. Normalized slip is the total distance a single leg travels backwards in the world frame divided by the forward distance traveled by the body. We present an average value for all four legs. Higher values of $\bar{s}$ indicate increased backwards motion of the legs, decreased propulsion, and reduced performance. The average value of $\bar{s}$ is $0.10 \pm 0.06$ ($n=9$) for the optimized trajectories, which is expected since we demand high performance from the robot. The closed-loop experimental trajectories slipped slightly more, with an average $\bar{s}$ of $0.24 \pm 0.14$ ($n=9$), and is one of the factors that could have resulted in decreased performance. In addition,  the optimizer also finds an intuitive jumping trajectory where all four legs first build spring potential energy, and then simultaneously push into the ground. 

\subsection{Limitations} 

Due to the nonconvexity of the trajectory optimization problem,  the optimizer can get stuck in poor local optima depending on the initial values of the decision variables. This is most clear at \SI{30}{\hertz} stride frequencies, where laterally symmetric gaits have achieved higher per cycle velocities \cite{goldberg2017gait}. Furthermore, our MATLAB implementation requires several minutes to compute the optimized plans, and significant speed improvements could be made with a C++ implementation. Finally, unmodeled serial compliance alters the transmission kinematics at higher frequencies, decreasing the utility of the optimized trajectories and the robot's performance.
 

\section{Conclusions and Future Work}
\label{sec:conclusions}
In this paper, we develop and evaluate a framework for modeling, planning, and controlling dynamic behaviors for legged microrobots. We develop a complete dynamic model of a complex microrobot, and adapt a state-of-the-art contact-implicit trajectory optimization algorithm to generate physically accurate whole-body locomotion plans for a variety of operating conditions. These locomotion plans are executed on the robot and result in improved performance.

Future work can improve upon both the existing optimization framework and the realization of optimized trajectories on the robot. For microrobots with high frequency passive dynamics, the contact mode trajectory often changes over longer timescales than the state and input trajectories. Encoding this into the NLP formulation by using higher order time-stepping methods could aid convergence and avoid unnecessary contacts without sacrificing richness in the contact force trajectories. This would allow for planning over longer horizons and enable optimization of a broader class of aperiodic behaviors. Furthermore, implementing whole-body locomotion controllers and/or learning-based methods to compensate for unmodeled effects would improve translation to the physical system, resulting in increasingly dynamic behaviors.

\section*{Acknowledgements}

This work is partially funded by the Wyss Institute for Biologically Inspired Engineering, the National Defense Science and Engineering Graduate Fellowship, and the National Science Foundation (Grant Number IIS-1657186). Any opinion, findings, and conclusions or recommendations
expressed in this material are those of the authors and do
not necessarily reflect the views of the National Science
Foundation. In addition, the prototypes were enabled by equipment supported by the ARO DURIP program (award $\#$W911NF-13-1-0311).

\bibliographystyle{plainnat}
\bibliography{FrictionTrajOpt_archive.bbl}
\end{document}